\documentclass[letterpaper]{article} 
\usepackage{aaai25}  
\usepackage{times}  
\usepackage{helvet}  
\usepackage{courier}  
\usepackage[hyphens]{url}  
\usepackage{graphicx} 
\def\UrlFont{\rm}  
\usepackage{natbib}  
\usepackage{caption} 
\usepackage{amsmath}
\usepackage{algorithm}
\usepackage{algorithmic}
\usepackage{booktabs}
\usepackage{bm}

\usepackage{newfloat}
\usepackage{listings}
\DeclareCaptionStyle{ruled}{labelfont=normalfont,labelsep=colon,strut=off} 
\floatstyle{ruled}
\newfloat{listing}{tb}{lst}{}
\floatname{listing}{Listing}
\usepackage{subfig}

\usepackage{xcolor}

\newcommand{\zw}[1]{#1}

\title{BigMac: A Communication-Efficient Mixture-of-Experts Model Structure for Fast Training and Inference}
\author {
    Zewen Jin\textsuperscript{\rm 1 2}\footnotemark[1],
    Shengnan Wang\textsuperscript{\rm 2}\footnotemark[1],
    Jiaan Zhu\textsuperscript{\rm 1 3},
    Hongrui Zhan\textsuperscript{\rm 1},\\
    Youhui Bai\textsuperscript{\rm 2},
    Lin Zhang\textsuperscript{\rm 2},
    Zhenyu Ming\textsuperscript{\rm 2},
    Cheng Li\textsuperscript{\rm 1 3}
}
\affiliations {
    \textsuperscript{\rm 1}University of Science and Technology of China\\
    \textsuperscript{\rm 2}Huawei Technologies\\
    \textsuperscript{\rm 3}Institute of Artificial Intelligence, Hefei Comprehensive National Science Center\\
    zevin@mail.ustc.edu.cn, 
    wangshengnan12@huawei.com,
    andyzhu@mail.ustc.edu.cn,
    zhr2001@mail.ustc.edu.cn,
    baiyouhui@huawei.com,
    zhang.lin4@huawei.com,
    mingzhenyu1@huawei.com,
    chengli7@ustc.edu.cn
}

\usepackage{bibentry}
\begin{document}

\maketitle
\renewcommand{\thefootnote}{\fnsymbol{footnote}}

\footnotetext[1]{Zewen and Shengnan equally contributed to this work. }

\begin{abstract}
The Mixture-of-Experts (MoE) structure 
scales the Transformer-based large language models (LLMs) and improves their 
performance with only the sub-linear increase in computation resources. Recently, a fine-grained DeepSeekMoE structure is proposed, which can further improve the computing efficiency of MoE without performance degradation.
However, the All-to-All communication introduced by MoE has become a bottleneck, especially for the fine-grained structure, which typically involves and activates more experts, hence contributing to heavier communication overhead.

In this paper, we propose a novel MoE structure named BigMac, which is also fine-grained but with high communication efficiency. 
The innovation of BigMac is mainly due to that we abandon the \textbf{c}ommunicate-\textbf{d}escend-\textbf{a}scend-\textbf{c}ommunicate (CDAC) manner used by fine-grained MoE, which leads to the All-to-All communication  always taking place at the highest dimension. Instead, BigMac designs an efficient \textbf{d}escend-\textbf{c}ommunicate-\textbf{c}ommunicate-\textbf{a}scend (DCCA) manner. 
Specifically, we add a descending and ascending projection at the entrance and exit of the expert, respectively, which enables the communication to perform at a very low dimension. Furthermore, to adapt to DCCA, we re-design the structure of small experts, ensuring that the expert in BigMac has enough complexity to address tokens. 
Experimental results show that BigMac achieves comparable or even better model quality than fine-grained MoEs  with the same number of experts and a similar number of total parameters. 
\zw{
Equally importantly, 
BigMac reduces the end-to-end latency by up to 3.09$\times$ for training and increases the throughput by up to 3.11$\times$ for inference on state-of-the-art AI computing frameworks including Megatron, Tutel, and DeepSpeed-Inference.
}

\end{abstract}

%

\section{Introduction}
\label{sec:introduction}

Increasing the size of Transformer-based large language models (LLMs) can continuously improve downstream application performance.
Such a phenomenon, known as the scaling law, has been demonstrated by the auto-regressive dense models such as GPT-series~\cite{openai2024gpt4technicalreport} and Llama-series~\cite{dubey2024llama3herdmodels}. 
However, this comes at the price of higher computing complexity and more resource consumption. Fortunately, the Mixture-of-Experts (MoE) technique, capable of expanding the model size tens or even hundreds of times without significantly increasing the computation, is widely used in various emerging huge models, such as GShard~\cite{lepikhin2020gshard}, GLaM~\cite{du2022glam}, Switch Transformer~\cite{fedus2022switch}, and Mixtral~\cite{jiang2024mixtral}, each of which consists of hundreds of billion parameters or even beyond. 

\begin{figure}[!t]
    \centering
    \includegraphics[width = 0.88\columnwidth]
    {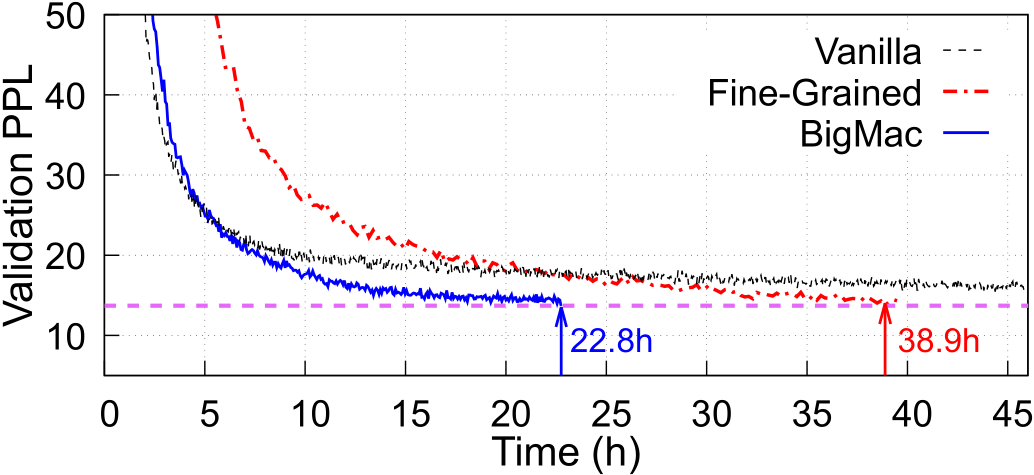}
    \caption{Convergence result comparison of MoE models with three structures. GPT-Fine-Grained takes 38.9 hours to reach the target perplexity of 13.69, while GPT-BigMac spends only 22.8 hours (1.7$\times$ faster). GPT-Vanilla fails to converge to the target perplexity under time budget.}
    \label{fig.loss.curve}
\end{figure}

Recently, DeepSeekMoE~\cite{dai2024deepseekmoeultimateexpertspecialization}, a fine-grained and more parameter-efficient MoE structure, has been proposed. Compared to conventional MoE models, for the same model size, DeepSeekMoE has significantly more experts per MoE layer and fewer parameters per expert. 
It was demonstrated that such a new structure can achieve comparable or even better results than conventional MoE models with much less time complexity and hence it is adopted by many later released models, such as Qwen2~\cite{yang2024qwen2} and DeepSeek-v2~\cite{deepseekai2024deepseekv2strongeconomicalefficient}. 

\begin{figure*}[!t]
    \centering
    \includegraphics[width=0.99\textwidth]{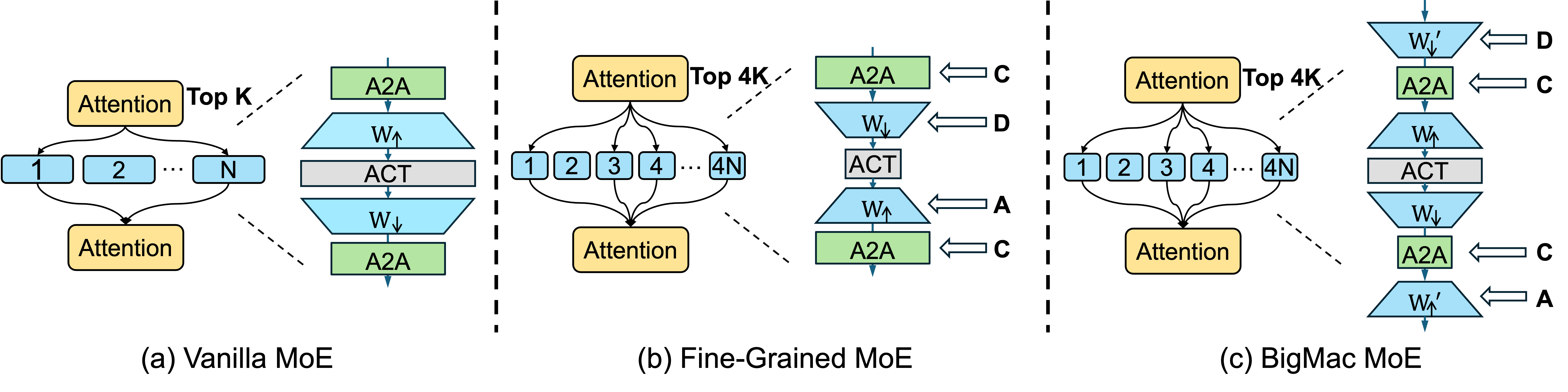}
    \caption{The MoE layers of different structures. Here, N represents the number of experts in the Vanilla MoE model and ACT represents the activation function like ReLU. $W_{\downarrow}$ and $W_{\uparrow}$ represent the descending and ascending projection matrix of an expert, respectively. $W_{\downarrow}'$ and $W_{\uparrow}'$ represent the projection matrices introduced in BigMac.}
    \label{fig.arch}
\end{figure*}

However, the MoE faces a serious All-to-All communication bottleneck during both training and inference. This is mainly because of the underlying expert parallelism (EP) strategy, which assigns experts to several hardware accelerators to avoid out-of-memory errors or improve efficiency~\cite{fedus2022switch}. As a result, for each MoE layer, two All-to-All communication steps are introduced for dispatching tokens to their best-fit experts, which may be stored in the other accelerators and then gathering the results back before proceeding to the next layer. 
Recent studies have already found that the All-to-All communication is a key issue leading to the low efficiency of MoE model training and inference~\cite{hu2024characterization}. Even worse, this bottleneck will be more pronounced in the fine-grained MoE structure, 
which generally requires to activate more experts to ensure performance. 

In this paper, we propose a novel efficient MoE structure, named BigMac, which is also fine-grained but can greatly reduce the All-to-All communication overhead. Note that existing fine-grained MoE models adopt the \textbf{c}ommunicate-\textbf{d}escend-\textbf{a}scend-\textbf{c}ommunicate (CDAC) manner, performing the All-to-All at a high dimension, leading to a heavy communication overhead. In contrast, BigMac designs an efficient \textbf{d}escend-\textbf{c}ommunicate-\textbf{c}ommunicate-\textbf{a}scend (DCCA) manner, which is capable of  performing the  All-to-All communication at a very low dimension.
Furthermore, to adapt to DCCA, we further re-design the structure of the small expert in BigMac, ensuring the complexity of each single expert for the overall performance of the whole model.
Specifically, each expert in BigMac is composed of an ascending projection and a descending projection with an activation in between, which is the opposite of fine-grained MoE models.

Briefly, we make the following contributions:
\begin{itemize}
\item We propose a novel MoE structure named BigMac, which can greatly improve the efficiency of MoE models for both training and inference. In addition, BigMac no longer suffers from problems such as the limited expert capacity and limited $top\_k$, which are common restrictions in the existing MoE structures.

\item We design a \textbf{d}escend-\textbf{c}ommunicate-\textbf{c}ommunicate-\textbf{a}scend (DCCA) strategy to ensure that the communication is always executed at the very low dimension, hence the communication overhead is greatly reduced. To guarantee the computing efficiency, BigMac adopts the idea of fine-grained MoE models, namely, each MoE layer is composed of a large number of small experts, while the structure of each small expert is re-designed to adapt to the DCCA strategy.

\item We pre-train the MoE models with different MoE structures and show that  BigMac converges faster than the other structures (shown in Figure~\ref{fig.loss.curve}). The results on multiple downstream tasks show that BigMac can achieve comparable or better performance against other baselines using the same amount of resources. 
Moreover, evaluations on state-of-the-art distributed training / inference frameworks, including Megatron, Tutel, 
\zw{
and DeepSpeed-Inference}, show that BigMac can significantly mitigate the communication overhead, reducing the end-to-end latency by up to 3.09$\times$ for training and increasing the throughput by up to 3.11$\times$ for inference.
\end{itemize}

\section{Related Work and Motivation}
\label{sec:related}

\noindent\textbf{Fine-grained MoE.} 
Starting from GShard~\cite{lepikhin2020gshard}, the Mixture-of-Experts (MoE) technology has been applied to the Transformer architecture, allowing for a significant increase in the number of parameters with only a sub-linear increase in computational resources. 
As illustrated in Figure~\ref{fig.arch}a, the dense Feed-Forward Network (FFN) modules in Transformer are replaced with MoE sub-layers, each consisting of multiple parallel experts. 
The number of experts activated by each token is defined as $top\_k$. 
However, it is challenging for these conventional MoE models to exploit expert specialization, since they are based on the top-1 or top-2 routing strategies with the coarse-grained expert activation. 
To address this issue, a new fine-grained MoE architecture is proposed in DeepSeekMoE~\cite{dai2024deepseekmoeultimateexpertspecialization}. 
To improve expert specialization, DeepSeekMoE maintains the same number of parameters as the conventional MoE models, while splitting experts into finer granularity and choosing a higher $top\_k$ for token distribution. 
Such an architecture with a large number of smaller experts has been adopted by DeepSeek-V2~\cite{deepseekai2024deepseekv2strongeconomicalefficient} and Qwen2-57B-A14B~\cite{yang2024qwen2}, demonstrating better model quality and computation efficiency than the conventional ones with a small number of large experts. 

Nevertheless, this fine-grained MoE architecture faces a severe communication problem for its training and inference tasks, due to the following reasons. 
First of all, to improve computation efficiency and cope with the single hardware accelerator's memory limit, the common practice is to leverage Expert Parallelism (EP) for assigning experts to different accelerators~\cite{fedus2022switch}. 
Second, EP requires injecting two costly All-to-All communication operations per MoE layer for distributing tokens to various experts and also gathering results for proceeding the computation to the next layer (green boxes in Figure~\ref{fig.arch}b). 
Third, the All-to-All communication already accounts for a large portion of the overall training or inference time, while its overhead as well as time ratio increases with the $top\_k$ value. 
Table~\ref{table.comm.ratio} shows the All-to-All time costs and time ratios of a fine-grained MoE model with 64 experts per layer and various $top\_k$ choices for training and inference jobs. 
When $top\_k$ is 1, the All-to-All overhead respectively contributes to 59.9\% and 51.2\% of the end-to-end time in training and inference. 
However, when $top\_k$ is 8, the All-to-All duration increases by 7.1$\times$ and 7.3$\times$, almost dominating the entire training and inference tasks, with proportions increasing to an astonishing 91.8\% and 90.6\%, respectively. 
In conclusion, considering that in the future new MoE models, it is very likely that their experts will become smaller and more numerous and the value of $top\_k$ will be larger, the optimization of All-to-All communication becomes urgent.

\begin{table}[!t]
    \centering
    \begin{tabular}{c|cc|cc}
    \toprule[0.8pt]
    \textbf{Expert Config} & \multicolumn{2}{c|}{\textbf{Training (ms)}} & \multicolumn{2}{c}{\textbf{Inference (ms)}} \\
    \textbf{$\bm{top\_k}$/\#experts } & \textbf{A2A}        & \textbf{Ratio}        & \textbf{A2A}        & \textbf{Ratio}        \\ \midrule[0.8pt]
    1 / 64   &     336.9          &  59.9\%            &     94.9			&	51.2\%    \\
    2 / 64   &     535.6          &  71.3\%            &     132.9			&	65.2\%        \\
    4 / 64   &     1,089.5          &  84.1\%            &     268.7			&	79.2\%    \\
    6 / 64   &     1,692.8          &  89.1\%            &     457.0			&	86.5\%         \\
    8 / 64   &     2,383.4          &  91.8\%            &     696.5			&	90.6\%    \\ 
    \bottomrule[0.8pt]
    \end{tabular}
    \caption{The All-to-All latency and its ratio in training and inference task of MoE models with small experts across various $top\_k$. The evaluation is conducted under the expert parallelism degree as 32 on 32 devices. The All-to-All duration increases by 7.1x and 7.3x when $top\_k$ is 8, corresponding to proportions of 91.8\% and 90.6\%.
}
    \label{table.comm.ratio}
\end{table}

\noindent\textbf{System-wise Optimizations. }
Fortunately, there have been a few initial attempts in the systems community to improve the scheduling of EP-enabled parallel training or inference. 
For instance, Lina~\cite{li2023accelerating} leverages a fine-grained scheduling strategy to avoid bandwidth contention between All-to-All communication and All-Reduce communication. 
Tutel~\cite{hwang2023tutel} schedules transmission jobs in a network topology-aware fashion 
to make full use of the intra-node and inter-node network bandwidth. 
Furthermore, FasterMoE~\cite{he2022fastermoe} and Tutel partition the input tokens into small chunks and overlap All-to-All communication with FFN computation in each expert. 
However, these efforts are designed for conventional MoE models, and when acting on fine-grained ones, their effect will be quite limited. This is mainly due to the fact that the amount of computation per expert is drastically reduced in the fine-grained MoE model, yet the amount of communication dominates the entire pipeline, and thus the space for bandwidth optimization and overlap optimization becomes very small.

\noindent\textbf{Communication Volume Reduction.} 
To address the communication bottleneck that is difficult to resolve at the system level, some have begun advocating data compression techniques. 
For instance, ScheMoE~\cite{shi2024schemoe} applies the ZFP compression algorithm~\cite{zfp} to tokens before transmission and indicates that such a compression technique can significantly reduce the All-to-All communication overhead 
and accelerate MoE training. 
However, such lossy MoE structure-agnostic compression schemes can lead to a decline in model quality, making them unsuitable for downstream tasks with high precision requirements. Furthermore, the extra compression and decompression steps can bring non-negligible computational overhead. 
Therefore, there is an urgent need for a new fine-grained MoE architecture from an algorithmic perspective with the following advantages: 1) significantly reducing data volumes transferred in All-to-All communication; 2) maintaining the same model quality; and 3) avoiding extra computation overhead, comparing to the state-of-the-art fine-grained MoE structures, as well as some of the above optimizations.

\begin{table}[!t]
    \centering
    \begin{tabular}{c|c}
    \toprule[0.8pt]
    \textbf{Notation} & \textbf{Description} \\ \midrule[0.8pt]
    $b$        & global batch size \\ \hline 
    $s$        & sequence length \\ \hline 
    $h$        & hidden dimension \\ \hline 
    $h\_f$     & FFN intermediate hidden dimension \\ \hline 
    $e$        & number of experts \\ \hline 
    $top\_k$   & number of experts to route to \\ \hline 
    $f$        & expert capacity factor \\ \hline 
    $ep$       & expert parallelism degree \\ \hline 
    $tp$       & tensor parallelism degree \\ \hline 
    $r$        & downscaling factor \\ \bottomrule[0.8pt]
    \end{tabular}
    \caption{Description of the notations used in this paper.}
    \label{table.notation}
\end{table}
\section{BigMac: Communication-Efficient MoE Structure}
\label{sec:methodology}
In this paper, we propose BigMac, a novel MoE structure that eliminates the well-known All-to-All communication bottleneck. Note that BigMac builds atop the success of fine-grained MoE models such as DeepSeekMoE and Qwen, where it also assigns a large number of small experts for each MoE layer, as shown in Figure~\ref{fig.arch}c. However, beyond this similarity, BigMac has the following two main differences in structure that reflect its design rationales, compared to fine-grained ones. 
\begin{itemize}
    \item \textbf{Low-dimensional communication}: we scale down the input/output tokens of experts to decrease the hidden dimension of the tokens to transfer, which greatly reduces the All-to-All communication overhead.
    \item \textbf{Performance assurance}: to adapt to the decreased dimension of input/output tokens, 
    we have to re-design the structure of each expert, using reversed projections to avoid the expert parameter count decreasing synchronously with the dimension and to align with the fine-grained MoE in terms of the total parameter count, to avoid diminishing the model quality.
\end{itemize}

Below, we will detail the BigMac's design with necessary notations, summarized in Table~\ref{table.notation}.

\subsection{DCCA: Low-dimensional Communication Strategy}
BigMac's efficient communication strategy is motivated by the estimation of the All-to-All communication overhead in each MoE layer of the fine-grained MoE models. This overhead can be described as
\begin{equation}
C = 2 \times top\_k \times \frac{ep-1}{ep}bsh,    
\label{equation.comm}
\end{equation}
which is proportional to the standard hidden dimension $h$. 
For the fine-grained MoE model, as shown in Figure~\ref{fig.arch}b, the model follows a  \textbf{c}ommunicate-\textbf{d}escend-\textbf{a}scend-\textbf{c}ommunicate (\textbf{CDAC}) manner, namely, the dimension of the tokens will be scaled down by a descending projection after the first All-to-All communication, and further be scaled up before the second All-to-All communication. Therefore, actually the fine-grained MoE model always transmits the token at the highest dimension, contributing to the serious overhead analyzed previously. Inspired by this fact, we ask a key question: \textit{is it possible for models like fine-grained MoEs communicate at low-dimensional level while maintaining the overall performance without degradation?}

To this end, as shown in Figure~\ref{fig.arch}c, at each MoE layer, BigMac moves the descending and ascending projections outside of every small expert and places the descending projection before the first All-to-All operation for remarkably scaling down tokens sent to their best-fit experts. This change allows the communication to happen at the lowest dimension. Following this, we place the ascending projection after the second All-to-All operation to scale up the tokens to their standard sizes. 
In contrast to the above CDAC manner used in fine-grained MoE models, BigMac follows a \textbf{d}escend-\textbf{c}ommunicate-\textbf{c}ommunicate-\textbf{a}scend (\textbf{DCCA}) manner. 
Within the DCCA execution, the whole process of the MoE module is described by the following equation: 
\begin{align}
x' = xW_{\downarrow}' ; \hspace{0.3cm}
y' &= {\sum_{i \in T} p_i(x)E_i(x')} ; \hspace{0.3cm}
y = y'W_{\uparrow}' .
\label{equation.moe.bigmac}
\end{align}
Here, $x$ and $y$ represent the output and input of two consecutive attention layers, $W_{\downarrow}'$ and $W_{\uparrow}'$ are the descending and ascending projection matrices, $T$ refers to the set of $top\_k$ experts for token distribution, $E_i$ refers to the expert computation in BigMac, 
and $p_i$ refers to the gate-value of activating the $i_{th}$ expert. Note that we can choose to use either $x$ or $x'$ as the input of the gating function for token routing. Here, we choose $x$, the vector before downscaling for routing, since the routing function is computationally efficient and a high-dimensional input vector generally leads to more accurate routing. In conclusion, DCCA reduces $C$ in Equation~\ref{equation.comm} into a much smaller $C'$ by changing $h$ to $rh$, where $r$ is the downscaling factor. Later, we will explain the value assignment to $r$ and overall communication savings.

\subsection{BigMac Expert Design}
Based on the DCCA strategy, following the expert structure of the fine-grained MoE models is impractical. Otherwise, the expert will have much fewer parameters and consequently hurt model quality.
Recall that expert computation can be described as $E(x) = \sigma(xW_{h\times h\_f}) W_{h\_f\times h}$, where $\sigma$ is an activation function, $h$ is the dimension of the input/output tokens and $h\_f$ refers to the  intermediate dimension. Compared to CDAC, DCCA significantly reduces the input/output dimension $h$, resulting in a smaller $E(x)$ with the same intermediate dimension $h\_f$.

As a result, to align BigMac's model size to that of fine-grained MoE models, we should increase the dimension $h\_f$. The specific structure of the expert designed for adapting the DCCA strategy is shown in Figure~\ref{fig.arch}c. From the appearance,
it is closer to the conventional MoE structure in Figure~\ref{fig.arch}a, and it can be seen as swaping the two projection matrices of fine-grained MoE experts in Figure~\ref{fig.arch}b.
In this way, the expert in the fine-grained MoE in Equation~\ref{equation.expert.lightweight} can be replaced with the one in BigMac as shown in Equation~\ref{equation.expert.bigmac}:
\begin{align}
E_i(x)  &= \sigma(xW_{i, \downarrow}) W_{i, \uparrow} ,\label{equation.expert.lightweight} \\
E_i(x) &= \sigma(xW_{i, \uparrow}) W_{i, \downarrow}. \label{equation.expert.bigmac}
\end{align}
It can be verified that the BigMac expert  involves the same size and same computational complexity compared with the expert in fine-grained MoE.

\begin{table*}[!t]
    \centering
    \begin{tabular}{c|c|c}
\toprule[0.8pt]
    \textbf{Metrics}        & \textbf{GPT-Fine-Grained} & \textbf{GPT-BigMac} \\ \midrule[0.8pt]
    \#Param & $(4h^2+8h+(2rh^2+2rh)e)l+(v+e+2)h$            & $(4h^2+8h+(2rh^2+2rh)e)l+(v+e+2)h+\mathbf{2rlh^2}$          \\ \hline
    \#FLOPs & $12bslh^2(2+\frac{s}{h}+\frac{v}{2lh}+r top\_k)$            & $12bslh^2(2+\frac{s}{h}+\frac{v}{2lh}+r top\_k)+\mathbf{12rbslh^2}$          \\  \hline
    \#A2A      & $8bslhtop\_k\frac{ep-1}{ep}$            & $8bslhtop\_k\frac{ep-1}{ep}\mathbf{r}$          \\   \bottomrule[0.8pt]
    \end{tabular}
    \caption{\zw{Statistics of two MoE models. \#Param refers to the number of parameters, \#FLOPs refers to the number of floating-point operations of an iteration for different MoE structures, and \#A2A refers to the transfer size of All-to-All communication.}}
    \label{table.analysis.math}
\end{table*}

\begin{table}[!t]
    \centering
    \begin{tabular}{c|c|c}
\toprule[0.8pt]
    \textbf{Metrics}        & \textbf{GPT-Fine-Grained} & \textbf{GPT-BigMac} \\ \midrule[0.8pt]
    \#Param  & 3.73B            & 3.78B (+1.35\%)          \\ \hline
    \#FLOPs  & 3,490.67 T           & 3,649.00 T (+4.54\%)          \\  \hline
    \#A2A     & 1,488.00 GB            & 372.00 GB (-75.00\%)          \\   \bottomrule[0.8pt]
    \end{tabular}
    \caption{Statistics for two MoE models with BF16 precision and an expert parallelism degree of 32 on 32 devices.}
    \label{table.analysis.number}
\end{table}

\begin{table}[!t]
\centering
\begin{tabular}{l|c}
Hyper-Params & Values \\
\toprule[0.8pt]
\#Layers  $l$     &24\\
\#Heads  $a$     &16\\
Hidden Dimension $h$     &2,048\\
Sequence Length $s$     & 2,048              \\
Vocabulary Size $v$     & 50,257             \\
Global Batch Size $b$     & 0.5 M             \\
Dropout Rate & 0.1 \\
Expert Capacity Factor $f$     & 1.2             \\
Load Balance Type & aux\_loss \\
Balance Coefficient $\alpha$ & 0.001  \\
\midrule[0.6pt]
Optimizer &               Adam     \\
Adam $\epsilon$, $\beta$ &           1e-8, (0.9,0.95)               \\
Weight Decay & 0.1                             \\
Learning Rate     & 3.0e-4 \\
Minimum Learning Rate     & 3.0e-5  \\
Learning Decay Steps & 5,200 \\
Learning Rate Decay Style &  cosine   \\
Warmup Steps &    1,200  \\
Gradient Clipping & 1.0 \\
Random Seed     &        1,234  \\
\bottomrule[0.8pt]
\end{tabular}
\caption{\zw{Hyper-parameters of pre-training to compare the validation perplexity curves in Figure~\ref{fig.loss.curve}.}}
\label{table.model.setup.same.time}
\end{table}
\begin{table*}[!t]
\centering
\begin{tabular}{c|cccc|cccc}
\toprule[0.8pt]
\begin{tabular}[c]{@{}c@{}}MoE \\ Structure\end{tabular}& \begin{tabular}[c]{@{}c@{}}PTB\\ (PPL↓)\end{tabular} & \begin{tabular}[c]{@{}c@{}}WikiText-\\ 103 (PPL↓)\end{tabular} & \begin{tabular}[c]{@{}c@{}}WikiText2\\ (PPL↓)\end{tabular} & \begin{tabular}[c]{@{}c@{}}LAMBADA\\ (ACC↑)\end{tabular} & \begin{tabular}[c]{@{}c@{}}HellaSwag\\ (ACC↑)\end{tabular} & \begin{tabular}[c]{@{}c@{}}WinoGrande\\ (ACC↑)\end{tabular} & \begin{tabular}[c]{@{}c@{}}PIQA\\ (ACC↑)\end{tabular} & \begin{tabular}[c]{@{}c@{}}RACE-H\\ (ACC↑)\end{tabular} \\ \midrule[0.8pt]
Fine-Grained    & 51.0          & 18.2 & 16.8          & 39.9 & 31.6 & 50.7          & 65.1 & 30.5          \\
BigMac      & \textbf{34.9} & \textbf{16.8} & \textbf{15.8} & \textbf{40.8}          & \textbf{33.2}          & \textbf{51.1} & \textbf{65.2}          & \textbf{31.3} \\ \bottomrule[0.8pt]
\end{tabular}
\caption{\zw{Downstream results for different MoE models (based on GPT3-XL) after training with the same time.}}
\label{table.eval.tasks.same.time}
\end{table*}
\begin{table*}[!t]
\centering
\begin{tabular}{c|cccc|cccc}
\toprule[0.8pt]
\begin{tabular}[c]{@{}c@{}}MoE \\ Structure\end{tabular}& \begin{tabular}[c]{@{}c@{}}PTB\\ (PPL↓)\end{tabular} & \begin{tabular}[c]{@{}c@{}}WikiText-\\ 103 (PPL↓)\end{tabular} & \begin{tabular}[c]{@{}c@{}}WikiText2\\ (PPL↓)\end{tabular} & \begin{tabular}[c]{@{}c@{}}LAMBADA\\ (ACC↑)\end{tabular} & \begin{tabular}[c]{@{}c@{}}HellaSwag\\ (ACC↑)\end{tabular} & \begin{tabular}[c]{@{}c@{}}WinoGrande\\ (ACC↑)\end{tabular} & \begin{tabular}[c]{@{}c@{}}PIQA\\ (ACC↑)\end{tabular} & \begin{tabular}[c]{@{}c@{}}RACE-H\\ (ACC↑)\end{tabular} \\ \midrule[0.8pt]
Vanilla           & 57.6          & 22.3          & 20.1          & 33.8          & 28.7          & 50.3          & 61.1          & 29.5          \\
Fine-Grained    & 67.7          & \textbf{19.1} & 17.9          & \textbf{38.3} & \textbf{31.0} & 49.5          & \textbf{65.0} & 29.8          \\
BigMac      & \textbf{52.3} & \textbf{19.1} & \textbf{17.7} & 37.6          & 30.8          & \textbf{51.3} & 64.2          & \textbf{30.7} \\ \bottomrule[0.8pt]
\end{tabular}
\caption{Downstream results for different MoE models (based on GPT3-Medium) after training with the same number of tokens. }
\label{table.eval.tasks.same.step}
\end{table*}

\begin{table*}[!t]
  \centering
  \begin{tabular}{c|ccccccccc}
  \toprule[0.8pt]
    Depth&10\%&20\%&30\%&40\%&50\%&60\%&70\%&80\%&90\%\\
  \midrule[0.8pt]
    Fine-Grained&99.1&99.3&99.0&98.6&98.4&98.3&98.3&98.2&97.9\\
    BigMac&100.0&99.9&99.4&99.0&98.8&98.6&98.5&98.3&98.1\\
    \bottomrule[0.8pt]
  \end{tabular}
  \caption{\zw{Recall scores of NeedleInAHaystack for different MoE models after training with the same number of tokens. }}
  \label{table.eval.tasks.same.step.needle}
\end{table*}

\subsection{Advantages Beyond Efficient Communication}
Except communication efficiency, BigMac further possesses many beneficial characteristics.
\subsubsection{Enabling Dropless Token Routing.}
In both training and inference phases of MoE, the token routing imbalance problem  occurs frequently, and it results in a severe straggler problem.
To reduce the overhead brought by the  imbalanced routing, most of the existing MoE models will set a threshold for the expert capacity~\cite{fedus2022switch}, determined by the \textit{expert capacity factor} $f$, which is often  set in a range from 1 to 1.25. Each expert will drop the tokens exceeding the expert capacity. 
It was demonstrated in~\cite{sanseviero2023moe} that the quality of the MoE model can be continuously improved by increasing the capacity factor, which implies that token dropping is harmful for model's generation. 
To ensure the performance, the recently proposed DeepSeekMoE and Mixtral remove the expert capacity limit, at the cost of high communication overhead~\cite{dai2024deepseekmoeultimateexpertspecialization,jiang2024mixtral,xue2024openmoeearlyeffortopen}.
Fortunately, the communication overhead has been greatly mitigated in BigMac. The increased token transmission brought by removing expert capacity limit will not significantly affect the overall training or inference efficiency.

\subsubsection{Enabling Flexible Selection of $\mathbf{top\_k}$.}
The number of activated experts, $top\_k$, is another key factor affecting model quality and overall latency.
To some extent, a larger $top\_k$ contributes to better model performance~\cite{dai2024deepseekmoeultimateexpertspecialization}.
However, a larger $top\_k$ corresponds to a heavier communication overhead, leading to lower efficiency for training and inference. Taking this into account, the existing MoE models generally select a relatively small $top\_k$.
Considering the high efficiency of BigMac in both computation and communication, BigMac is able to withstand a much larger $top\_k$ to enhance the performance. Hence, BigMac provides a  more flexible choice for practitioners.

\subsection{Analysis of Different MoE Structures}
To understand how BigMac differs from the existing fine-grained MoE structure, we analyze the parameter size and the number of FLOPs as well as the communication overhead 
\zw{
for different MoE structures in Table~\ref{table.analysis.math}.  
It indicates that the two additional projection matrices in BigMac can significantly reduce the All-to-All transmission size by a ratio of $(1-r)$ while involving negligible overhead.
For a more intuitive elaboration, we show the concrete numbers with GPT3-XL as the base model in Table~\ref{table.analysis.number}, where we activate 8 experts out of 64 experts and set the downscaling factor $r$ as 0.25, considering a similar setting of DeepSeek-V2, i.e., scaling down from 5,120 to 1,536. 
}
Table~\ref{table.analysis.number} shows that the additional scaling matrices introduce only 4.54\% FLOPs while reducing up to 75\%  communication overhead.

\section{Pre-Training and Downstream Evaluation}
\label{sec:zero.shot.tasks}
\subsection{Pre-Training Tasks}
To show the acceleration of training convergence with constant model quality, we first pre-train three MoE models with different MoE structures, namely GPT-Vanilla, GPT-Fine-Grained, and GPT-BigMac, all of which use GPT3-XL as the base model. Vanilla represents the conventional MoE with large experts, Fine-Grained refers to the MoE model with small experts, while BigMac is our design. For a fair comparison, we keep the same parameter size of MoE layers across the three models. We use the Wikipedia dataset~\cite{wikidump} containing 3.6 B tokens to train these models on Megatron~\cite{megatron}, one of the state-of-the-art  LLM training frameworks.

Figure~\ref{fig.loss.curve} shows the curve of validation perplexity of pre-training, indicating that GPT-BigMac converges much faster than others and achieves the lowest validation perplexity within the same time. 
For example, to achieve the same validation perplexity of 13.69, GPT-Fine-Grained requires 38.9 hours while GPT-BigMac only needs 22.8 hours, which is 1.71$\times$ faster.
In addition, among the three model structures, GPT-Vanilla fails to converge to the same validation perplexity under the time budget, indicating that with the same parameter size, the MoE structure with small experts outperforms the conventional MoE.
Further, with the evaluation on WikiText2~\cite{merity2016pointer}, GPT-BigMac achieves the perplexity score of 17.4, while GPT-Vanilla and GPT-Fine-Grained get 27.4 and 17.9, respectively.
\zw{
The hyper-parameters for pre-training are shown in Table~\ref{table.model.setup.same.time} and the degree of Tensor Parallelism, Expert Parallelism, and Data Parallelism is set as 4, 4, and 2, respectively.
}
\subsection{Downstream Tasks}
\zw{
To demonstrate how BigMac impacts the model quality on downstream tasks, we utilized a larger dataset named OpenWebText2 dataset~\cite{openwebtext} with 14.8 B tokens. 
First, we compare the performance after training for the same duration (8 days) based on the hyper-parameters in Table~\ref{table.model.setup.same.time}.
}
We evaluate the fine-grained and BigMac variants, which are based on GPT3-XL, on eight popular zero-shot tasks, including four long-term dependence prediction tasks (LAMBADA~\cite{paperno2016lambada}, PTB~\cite{marcus1993building}, WikiText103 and WikiText2~\cite{merity2016pointer}) and four question answering tasks (PIQA~\cite{Bisk2020}, HellaSwag~\cite{zellers2019hellaswag} and WinoGrande~\cite{sakaguchi2019winogrande}, and RACE-H~\cite{lai-etal-2017-race}). 
\zw{
Table~\ref{table.eval.tasks.same.time} shows the results of the eight downstream tasks in terms of accuracy (ACC) and perplexity (PPL). 
It shows that after training with the same time and GPU resources, GPT-BigMac gives a better model quality.
}

\zw{
Next, we further compare the performance after training for the same number of steps and tokens (3 epochs for all models).
For efficiency, we use GPT3-Medium as the base model and use the same hyper-parameters in Table~\ref{table.model.setup.same.time}, except that the values of Hidden Dimension, Learning Decay Steps, and Warmup Steps are 1,024, 28,000, and 5,000, respectively.
}
Table~\ref{table.eval.tasks.same.step} shows the results of the eight downstream tasks. 
GPT-BigMac delivers comparable or better results against GPT-Fine-Grained, achieving the best performance for 5 out of 8 tasks. 
For example, GPT-BigMac surpasses GPT-Fine-Grained by a score of 0.9 on RACE-H. 
Both GPT-BigMac and GPT-Fine-Grained outperform GPT-Vanilla, which shows the superiority of fine-grained MoE models. 
\zw{
In addition, we also evaluate two tasks, including GovReport~\cite{govreport} for summarization and NeedleInAHaystack~\cite{needle} for retrieval. 
GPT-BigMac achieves the score of 19.5 for GovReport, which is better than 17.7 achieved by GPT-DeepSeek. 
For NeedleInAHaystack, GPT-BigMac delivers comparable recall scores across different depths (Table~\ref{table.eval.tasks.same.step.needle}). 
}
\section{Training and Inference Speedups}
In the last section, we have shown that compared with the traditional MoE structure, MoE structures with small experts are more powerful. In this section, we further compare the communication efficiency of the fine-grained MoE structure and BigMac in more depth.
\subsection{Experimental Setup} 
We intensively profile the time ratios of training and inference for GPT-Fine-Grained and GPT-BigMac,  based on the state-of-the-art frameworks Megatron~\cite{megatron.paper}\zw{, Tutel~\cite{hwang2023tutel}, and DeepSpeed-Inference~\cite{deepspeed.inference}. 
}
Megatron supports various parallelism strategies including data parallelism (DP), tensor parallelism (TP), and expert parallelism (EP). 
Tutel is a specialized framework to optimize the All-to-All communication for MoE models. 
\zw{
DeepSpeed-Inference supports techniques specialized for LLM inference including KV cache management to efficiently serve the models.
}
All the experiments are conducted on a cluster of 4 machines connected with 100 Gbps InfiniBand. Each machine has the same configuration and is equipped with eight  GPUs. Each GPU is connected with PCIe 4.0 x 16 and has 48 GB HBM, delivering up to 149.7 TFLOPS (FP16) with 96 cores. 
For all the experiments,  the input sequence length is 2,048 and the global batch size is 64. We mainly compare the two structures in terms of training step latency, the corresponding All-to-All latency, and the inference throughput.

\begin{figure}[!t]
    \centering
    \subfloat[Top8\label{fig.megatron.lat.e2e.top8}]{\includegraphics[height=0.3\columnwidth]{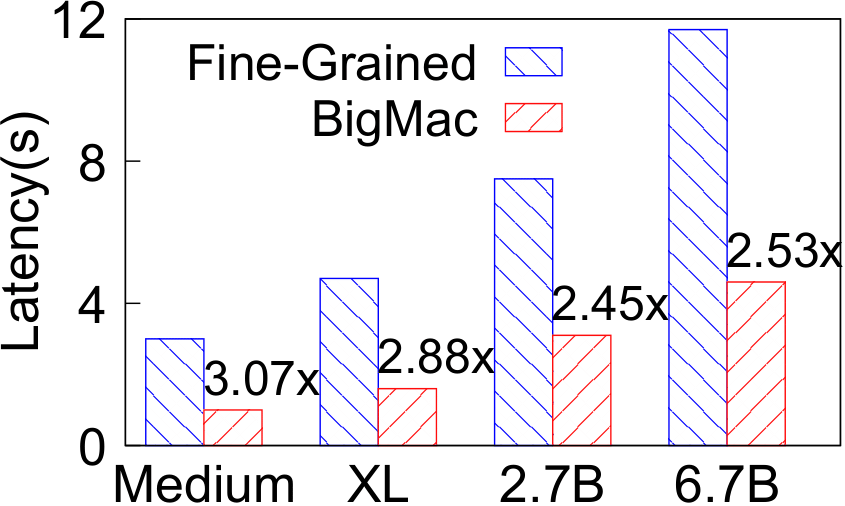}} 
    \hfill
    \subfloat[Top4\label{fig.megatron.lat.e2e.top4}]{\includegraphics[height=0.3\columnwidth]{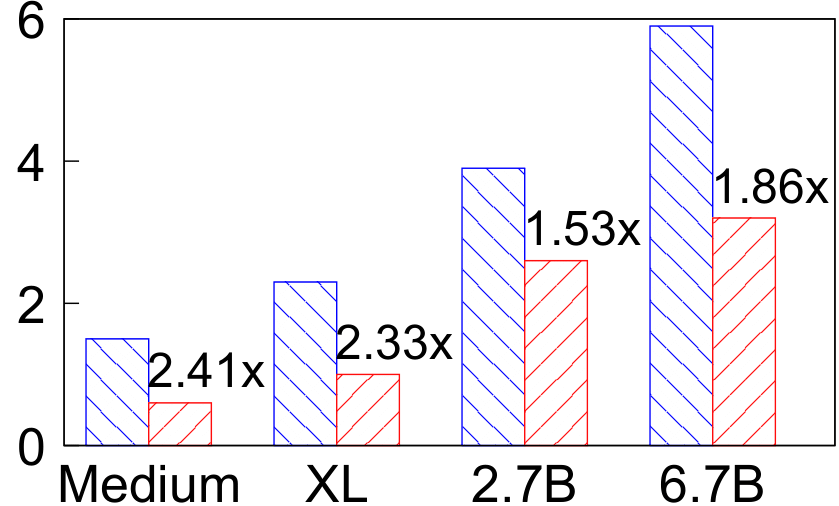}}
    \caption{Per-iteration training time comparison between the fine-grained structure and BigMac on Megatron. The models are constructed from four base models, namely GPT3-Medium, GPT3-XL, GPT3-2.7B, and GPT3-6.7B, ordered by the size of parameters. }
    \label{fig.megatron.latency.e2e}
\end{figure}

\subsection{Comparing Training Latency via  Megatron}
We first compare the training step time of fine-grained and BigMac models under the Megatron framework. Here, we adopt four base models including GPT3-Medium, GPT3-XL, GPT3-2.7B, and GPT3-6.7B.

Figure~\ref{fig.megatron.latency.e2e} shows that GPT-BigMac achieves the speedups of 1.53-2.41$\times$ and 2.45-3.07$\times$ than GPT-Fine-Grained for Top4 and Top8 routing settings, respectively. Note that larger $top\_k$ generally indicates the heavier communication, hence GPT-BigMac enjoys greater advantages in the Top8 setting. For the MoE models with small experts, larger $top\_k$ implies better performance to some extent. Due to the high communication efficiency, BigMac can choose a larger $top\_k$ than GPT-Fine-Grained.
Surprisingly,  GPT-BigMac using the Top8 routing can still outperform GPT-Fine-Grained using the Top4 routing by 27.7-55.4\% in terms of the end-to-end latency.

\begin{figure}[!t]
    \centering
    \subfloat[Top8\label{fig.train.breakdown.top8}]{\includegraphics[height=0.32\columnwidth]{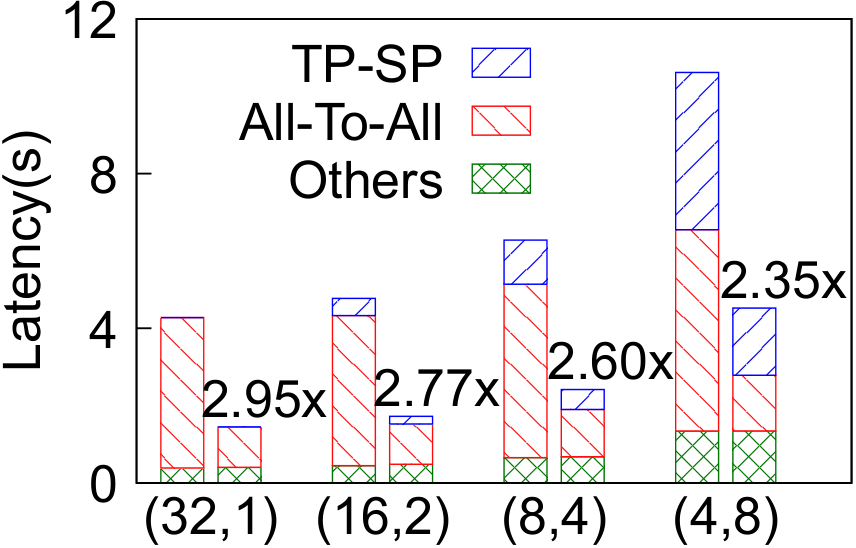}} 
    \hfill
    \subfloat[Top4\label{fig.train.breakdown.top4}]{\includegraphics[height=0.32\columnwidth]{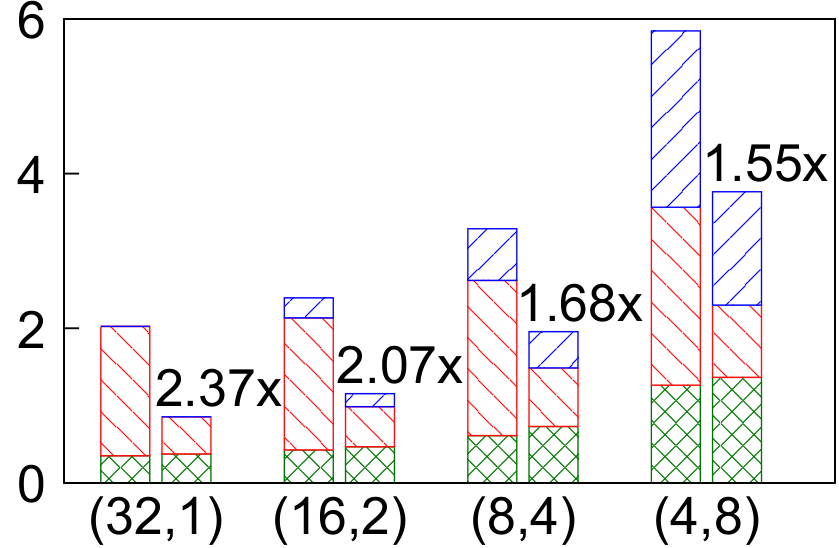}}
    \caption{Training time breakdown under different parallelism settings on Megatron. The labels \textit{($ep$, $tp$)} represent expert parallelism degree and tensor parallelism degree, respectively. For each group, the left bar is the result of GPT-Fine-Grained, and the right bar corresponds to GPT-BigMac. The numbers displayed on the right bar indicate the speedup in end-to-end latency.}
    \label{fig.megatron.latency.breakdown}
\end{figure}

\noindent\textbf{Breakdown Analysis.} 
To understand the above speedups in depth, we report the breakdown results for training with an emphasis on the All-to-All communication cost. 
In Figure~\ref{fig.megatron.latency.breakdown}, the (32, 1) groups refer to the setup with only the expert parallelism  and its degree $ep$ setting to  32. 
In this setting, BigMac achieves an end-to-end speedup of 2.37$\times$ and 2.95$\times$  under the Top4 and Top8 routing, respectively, compared to the fine-grained baseline, where the speedup w.r.t.  the All-to-All communication is 3.48$\times$ and 3.72$\times$, respectively.
In addition to the above pure expert parallelism setting, we also consider the combinations of various parallelism modes. We adopt tensor parallelism with the following settings. 
Specifically, we set the tensor parallelism degree $tp$ from 1 to 8,  and then adjust expert parallelism degree $ep$ by $ep=32/tp$.
In this situation,  BigMac can still reduce the All-to-All communication by 2.47-3.73$\times$ and the end-to-end latency by 1.55-2.77$\times$. 
\zw{
In Megatron, the TP-SP communication in the MoE layer involves the operations of All-to-All, All-Gather, and Reduce-Scatter within each TP group.
All these operations happen at the higher dimension in the original fine-grained structure and the lower dimension with the design of BigMac.
In this way, BigMac also reduces the TP-SP communication by 1.42-2.34$\times$ for different setups.
}
Finally, according to the results of the four parallelism settings shown in the figure, for the sake of efficiency, expert parallelism is preferred over tensor parallelism in our setting, as tensor parallelism involves more expensive all-reduce communication.

\begin{figure}[!t]
    \centering
    \subfloat[$ep$=32, Top8\label{fig.tpt.megatron.infer.ep32.top8}]{\includegraphics[height=0.29\columnwidth]{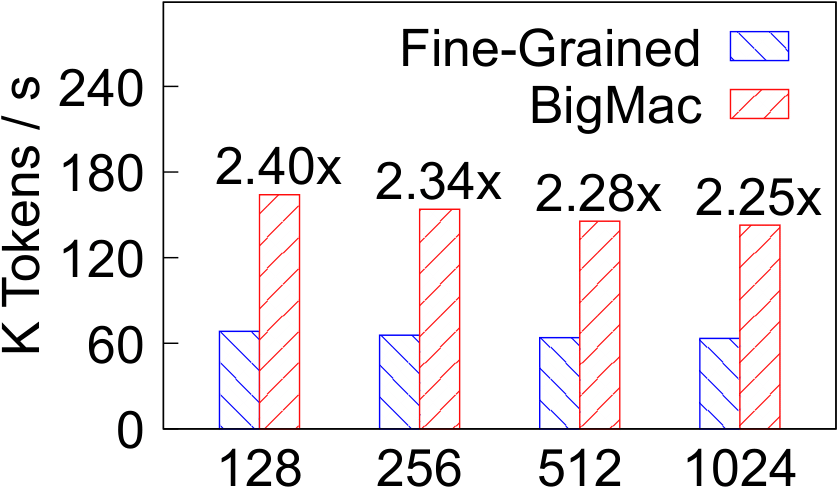}} 
    \subfloat[$ep$=32, Top4\label{fig.tpt.megatron.infer.ep32.top4}]{\includegraphics[height=0.29\columnwidth]{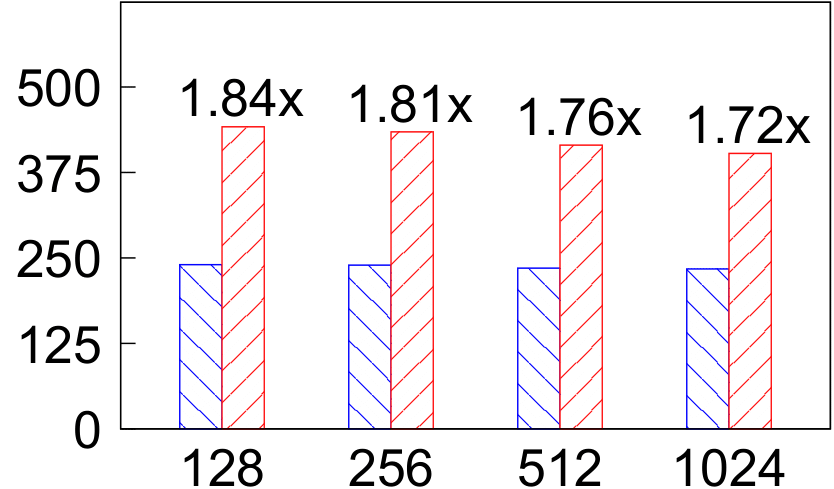}} 
    \hfill
    \subfloat[$ep$=16, Top8\label{fig.tpt.megatron.infer.ep16.top8}]{\includegraphics[height=0.29\columnwidth]{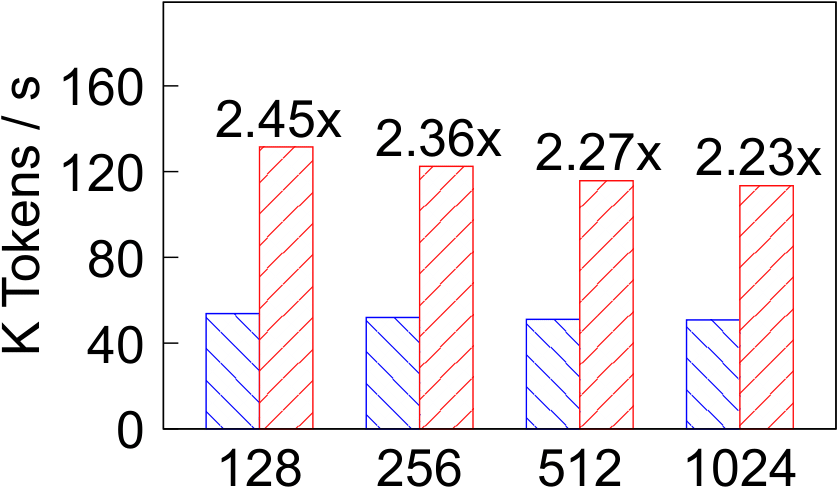}} 
    \subfloat[$ep$=16, Top4\label{fig.tpt.megatron.infer.ep16.top4}]{\includegraphics[height=0.29\columnwidth]{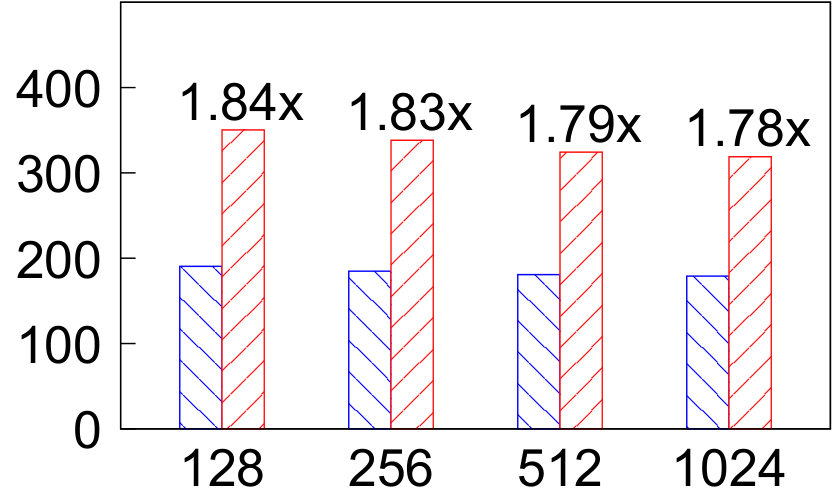}} 
    \caption{Inference throughput comparison between GPT-Fine-Grained and GPT-BigMac on Megatron. We conduct experiments with different numbers of GPUs with expert parallelism degree $ep$ and $top\_k$ values. The numbers under x-axis represents different prompt lengths.}
    \label{fig.tpt.megatron.infer}
\end{figure}
\subsection{Inference Throughput Comparison with Megatron}

For inference, we measure the throughput of the forward pass under the Megatron framework. We keep the number of the tokens per batch to be 128k, but with varying prompt lengths, ranging from 128 to 1,024. We use 16 and 32 GPUs for evaluation and we set the expert parallelism degree $ep$ to 16 and 32, respectively. Here we do not adopt tensor parallelism since it is less efficient.

Figure~\ref{fig.tpt.megatron.infer} shows that GPT-BigMac consistently outperforms GPT-Fine-Grained and achieves 1.72-2.45$\times$ speedups across all the settings. First, 
BigMac can obtain higher speedups with larger $top\_k$,. 
Second, the amplitude of speedup decreases slightly as the prompt length increases. 
Note that the larger prompt length brings heavier computation overhead in the attention layer, and then the proportion of All-to-All communication decreases correspondingly, especially for BigMac, which explains its slight decline in the inference throughput.

\subsection{Comparison on All-to-All Optimized System}
Finally, we investigate if BigMac's model structure can bring benefits further on systems which have already optimized the All-to-All bottleneck of MoE from systems perspectives. 
\zw{
For training, we evaluate on Tutel and for inference, we evaluate on Tutel and DeepSpeed-Inference.
}
We evaluate GPT-Fine-Grained and GPT-BigMac, using the GPT3-Medium as the base model, with different expert parallelism degrees and $top\_k$ values. 
In Tutel, we adopt the 2DH All-to-All communication technique and set the overlapping degree as 4 to hide communications with expert computations. 
In addition, Tutel supports dynamic capacity factor adaption, which avoids token dropping while reducing token padding. We measure with a fixed factor ($f$=1.2) and the dynamic capacity factor adaption ($f$=$\infty$), respectively.

\begin{figure}[!t]
    \centering
    \subfloat[Top8\label{fig.lat.tutel.ep32.top8}]{\includegraphics[height=0.32\columnwidth]{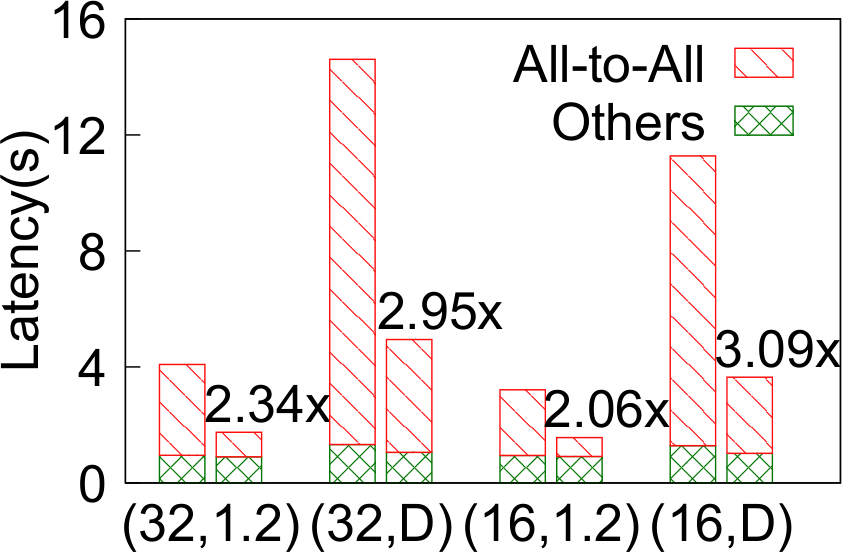}} 
    \hfill
    \subfloat[Top4\label{fig.lat.tutel.ep16.top8}]{\includegraphics[height=0.32\columnwidth]{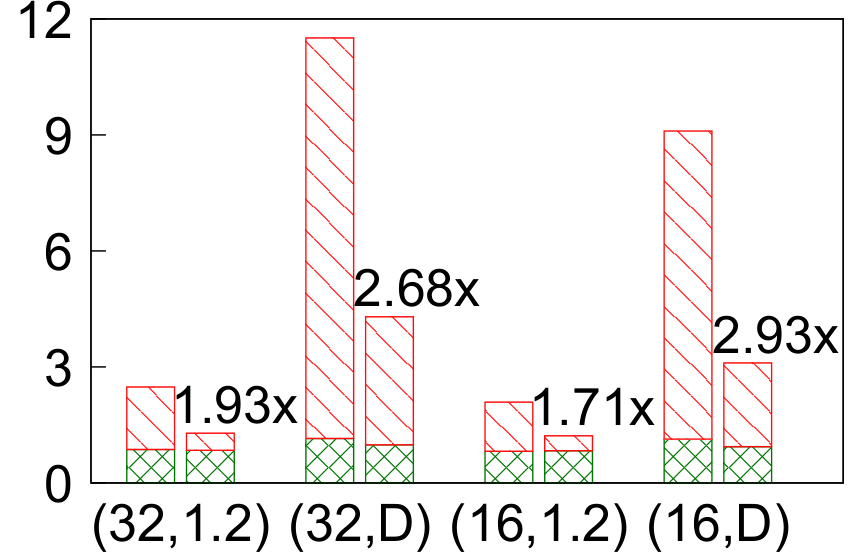}}
    \caption{Training time breakdown on Tutel. For each group, labels ($ep$, $f$) refer to the corresponding EP degree and the expert capacity factor $f$, where $f$=\textit{D} refers to the dynamic capacity factor adaption. For each group, the left bar is the result of GPT-Fine-Grained, and the right bar corresponds to GPT-BigMac.}
    \label{fig.tutel.latency}
\end{figure}
\noindent\textbf{Training Latency on Tutel.} 
Figure~\ref{fig.tutel.latency} shows the training speedups of GPT-BigMac, compared with  GPT-Fine-Grained under Top8/Top4 routing, and we show the results with fixed  capacity factor ($f$=1.2) and dynamic capacity factor ($f$=$\infty$), respectively. We can see that BigMac has significant speedups ranging from 1.71$\times$ to 3.09$\times$ in all the cases,  and BigMac shows greater advantages in Top8 routing and dynamic capacity setting, since both larger $top\_k$ and larger capacity indicate more data transmission.

\begin{figure}[!t]
    \centering
    \subfloat[$ep$=32, Top8\label{fig.tpt.tuel.infer.ep32.top8}]{\includegraphics[height=0.29\columnwidth]{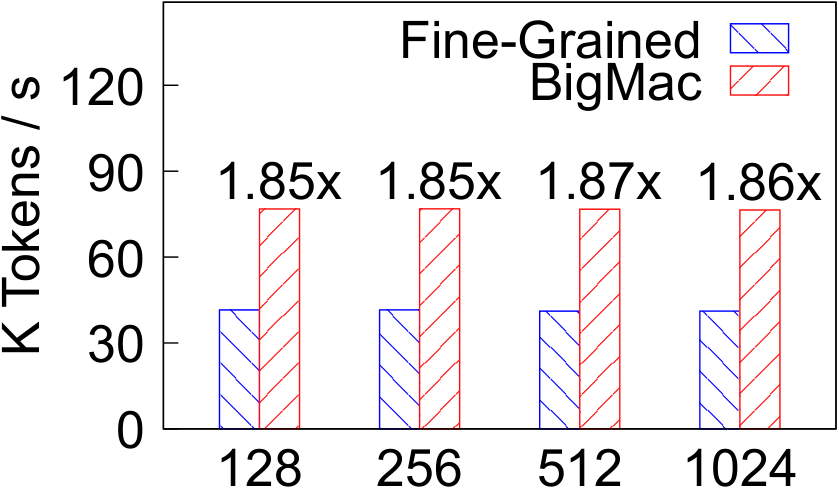}} 
    \subfloat[$ep$=32, Top4\label{fig.tpt.tuel.infer.ep32.top4}]{\includegraphics[height=0.29\columnwidth]{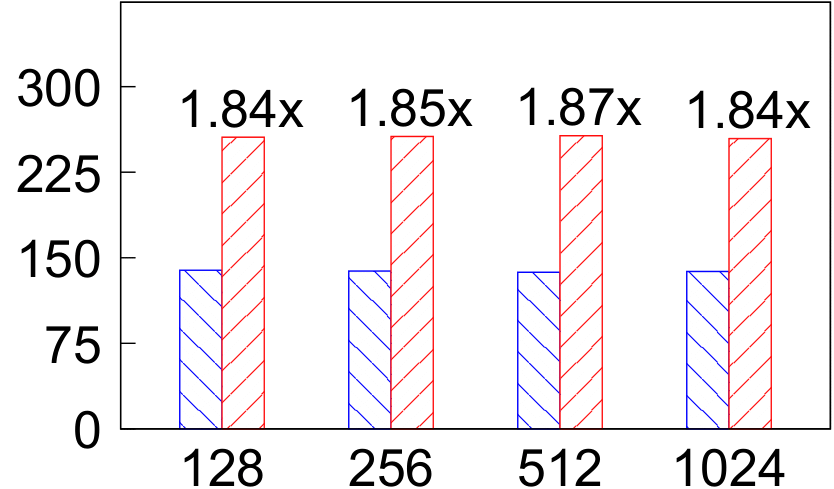}} \\
    \subfloat[$ep$=16, Top8\label{fig.tpt.tuel.infer.ep16.top8}]{\includegraphics[height=0.29\columnwidth]{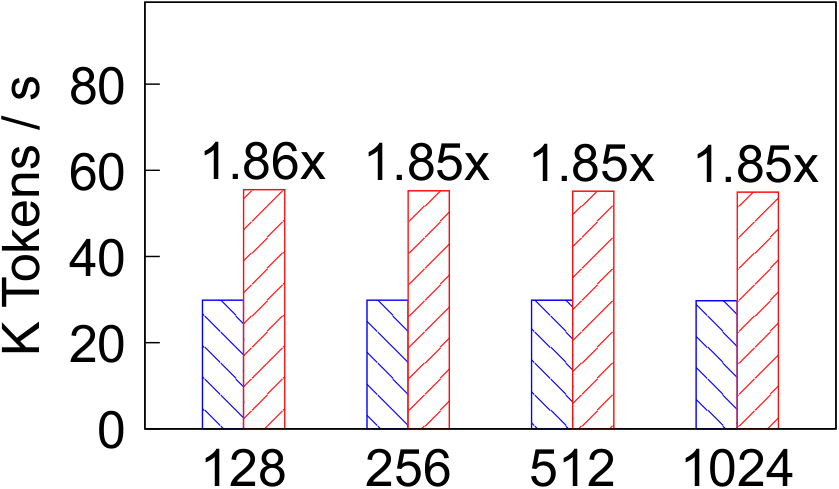}}  
    \subfloat[$ep$=16, Top4\label{fig.tpt.tuel.infer.ep16.top4}]{\includegraphics[height=0.29\columnwidth]{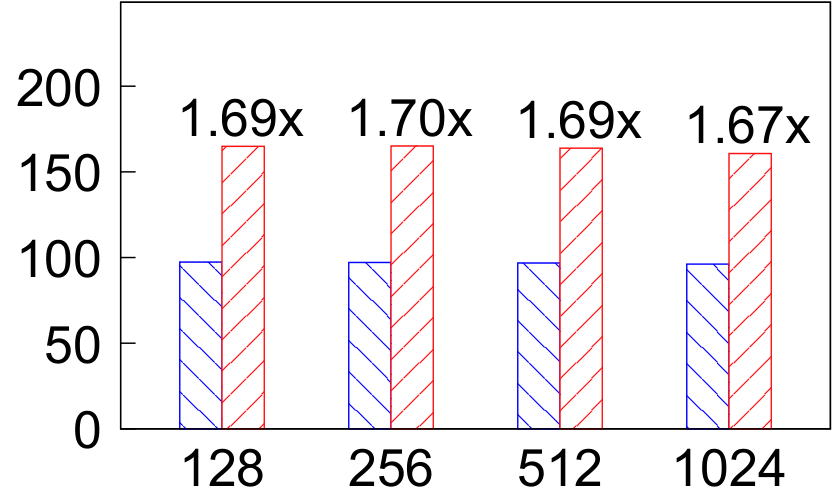}} 
    \caption{Inference throughput comparison between GPT-Fine-Grained and GPT-BigMac on Tutel with $f$=1.2. The numbers under x-axis represents different prompt lengths.}
    \label{fig.tpt.tutel.infer}
\end{figure}

\noindent\textbf{Inference Throughput on Tutel.} We summarize the inference throughput of GPT-Fine-Grained and GPT-BigMac on Tutel for different prompt lengths in Figure~\ref{fig.tpt.tutel.infer}. GPT-BigMac consistently outperforms GPT-Fine-Grained by 1.67-1.87$\times$, under different $top\_k$ value and expert parallelism degrees. This implies that with system optimizations enabled by Tutel, BigMac can still maintain a high throughput over different prompt lengths.

\zw{
\noindent\textbf{Inference Throughput on DeepSpeed-Inference.} 
We next compare the inference throughput of GPT-Fine-Grained and GPT-BigMac on DeepSpeed-Inference for different generation lengths. 
Table~\ref{table.tpt.speedup.deepspeed} shows the speedup of inference throughput under the prompt length of 128. 
The results show that on DeepSeepd-Inference, which involves techniques including KV cache management, GPT-BigMac consistently outperforms GPT-Fine-Grained by 1.62-3.11$\times$ over different generation lengths.
}

\begin{table}[!t]
  \centering
  \begin{tabular}{c|cccc}
  \toprule[0.8pt]
    Generation Length&1&2&5&10\\
  \midrule[0.8pt]
    $ep$=16,Top8&3.11$\times$&2.89$\times$&2.41$\times$&1.99$\times$\\
    $ep$=16,Top4&2.81$\times$&2.50$\times$&2.03$\times$&1.62$\times$\\
    \bottomrule[0.8pt]
  \end{tabular}
  \caption{\zw{Inference throughput speedup of GPT-BigMac on DeepSpeed-Inference under different generation lengths.}}
  \label{table.tpt.speedup.deepspeed}
\end{table}

\section{Discussion}
\label{sec:Discussion}

In Figure~\ref{fig.arch}c, BigMac introduces two additional scaling projections. However, the computation brought by the two projections is negligible compared with the benefits from the All-to-All communication reduction. 
For small models without the necessity of expert parallelism, BigMac indeed slightly increases the overall latency since no All-to-All communication is required in this case.
Therefore, BigMac is more suitable for large models which are the current trend of novel models. In our structure, the downscaling factor $r$ affects both the All-to-All communication overhead and the model quality. In this paper, for a fair comparison, we set the factor $r$ as 0.25 to ensure that the MoE models with three different structures involve similar number of parameters. One can adjust the ratio in real applications, according to the actual demand.

\section{Conclusion}
\label{sec:conclusion}
We proposed a novel MoE structure named BigMac which uses a \textbf{d}escend-\textbf{c}ommunicate-\textbf{c}ommunicate-\textbf{a}scend (DCCA) strategy to reduce the communication overhead by performing All-to-All operations at the lowest dimension. Results demonstrate that BigMac achieves comparable or superior model quality to the existing MoE structures, with significant speedups in training and inference across different platforms, making it a strong contender among MoE-based large language models.

\section*{Acknowledgements}
We thank the anonymous reviewers for their insightful comments. 
This work is supported by 
the Strategic Priority Research Program of the Chinese Academy of Sciences, Grant No. XDB0660101, XDB0660000, and XDB0660100. 
We thank the technical support from Huawei and computing resources from Institute of Artificial Intelligence, Hefei Comprehensive National Science Center.
Cheng Li is the corresponding author.

\bibliography{reference}

\end{document}